\documentclass[10pt,twocolumn,letterpaper]{article}

\usepackage[pagenumbers]{cvpr} % To force page numbers, e.g. for an arXiv version

\usepackage{booktabs} % 用于专业表格横线
\usepackage{makecell} % 用于单元格换行
\usepackage{graphicx} % 用于表格缩放
\usepackage[table]{xcolor}
\usepackage{comment}
\usepackage{bm}
\definecolor{lightgray}{gray}{0.9}

\usepackage{amsmath}
\usepackage{algorithm}
\usepackage{algcompatible}
\usepackage[noend]{algpseudocode}

\definecolor{cvprblue}{rgb}{0.21,0.49,0.74}
\usepackage[table]{xcolor} 
\usepackage{booktabs}
\usepackage{siunitx}
\usepackage[pagebackref,breaklinks,colorlinks,allcolors=cvprblue]{hyperref}

\def\paperID{5786} % *** Enter the Paper ID here
\def\confName{CVPR}
\def\confYear{2026}

\title{VideoSSM: Autoregressive Long Video Generation  with \\ Hybrid State-Space Memory}

\author{
    \textbf{Yifei Yu}$^{1,\dagger,*}$, \quad
    \textbf{Xiaoshan Wu}$^{1,*}$, \quad
    \textbf{Xinting Hu}$^{1}$, \quad
    \textbf{Tao Hu}$^{2, \P}$, \quad
    \textbf{Yang-Tian Sun}$^{1}$, \\
    \textbf{Xiaoyang Lyu}$^{1}$, \quad
    \textbf{Bo Wang}$^{1}$, \quad
    \textbf{Lin Ma}$^{2}$, \quad
    \textbf{Yuewen Ma}$^{2,\ddagger}$, \quad
    \textbf{Zhongrui Wang}$^{3,\ddagger}$, \quad
    \textbf{Xiaojuan Qi}$^{1,\ddagger}$ \\
    $^1$HKU \quad
    $^2$PICO, ByteDance \quad
    $^3$SUSTech \\
    $^{\dagger}$ Work done during internship at PICO \quad
    $^{*}$ Equal contribution \\
    $^{\P}$ Project lead \quad
    $^{\ddagger}$ Corresponding author \\
}

\begin{document}
% \maketitle
% \input{sec/0_abstract}    
% \input{sec/1_intro}
% \input{sec/2_formatting}
% \input{sec/3_finalcopy}

\twocolumn[{
\renewcommand\twocolumn[1][]{#1}
\maketitle

\begin{center}
    \centering
    \captionsetup{type=figure}
    \vspace{-4mm}
    \includegraphics[width=1\linewidth, trim=0 0 0 0, clip]{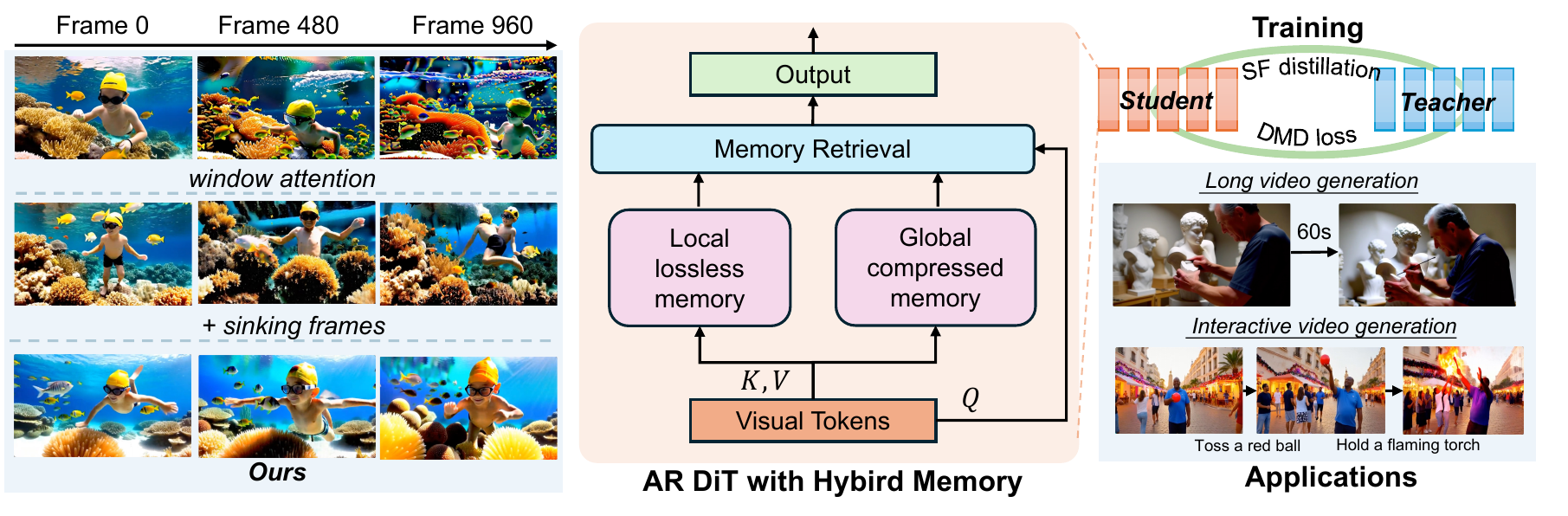}
    \vspace{-6mm}
    % \captionof{figure}{We introduce \textbf{VideoSSM}, an autoregressive video generation model with capable of generating , generating view-consistent stereo videos with intrinsic geometric understanding. StereoWorld can be applied to downstream tasks like VR/AR visualization as well as action planning in embodied intelligence.}
    \captionof{figure}{We introduce \textbf{VideoSSM}, an AR video diffusion model equipped with a novel hybrid memory architecture that combines a causal sliding-window local lossless cache with an SSM-based global compressed memory. Compared with prior AR Diffusion Transformers (AR DiTs) that use only window attention, which suffer from quality degradation and temporal drifting, or add sink frames, which reduce drift but cause content repetition and a lack of dynamism, our hybrid memory yields videos that remain both long-term consistent and progressively dynamic. Trained via Self Forcing distillation~\cite{huang2025self} via DMD loss~\cite{yin2024onestep} from a bidirectional teacher, VideoSSM supports highly stable long video generation and adaptive, interactive prompt-based video generation.}
\label{fig:teaser}
\end{center}
}]

\begin{abstract}
Autoregressive (AR) diffusion enables streaming, interactive long-video generation by producing frames causally, yet maintaining coherence over minute-scale horizons remains challenging due to accumulated errors, motion drift, and content repetition. We approach this problem from a memory perspective, treating video synthesis as a recurrent dynamical process that requires coordinated short- and long-term context. We propose VideoSSM, a Long Video Model that unifies AR diffusion with a hybrid state-space memory. The state-space model (SSM) serves as an evolving global memory of scene dynamics across the entire sequence, while a context window provides local memory for motion cues and fine details. This hybrid design preserves global consistency without frozen, repetitive patterns, supports prompt-adaptive interaction, and scales in linear time with sequence length. Experiments on short- and long-range benchmarks demonstrate state-of-the-art temporal consistency and motion stability among autoregressive video generator especially at minute-scale horizons, enabling content diversity and interactive prompt-based control, thereby establishing a scalable, memory-aware framework for long video generation. 
\end{abstract}
\section{Introduction}
\label{sec:intro}
% background for long video generation
Long-video generation is a longstanding goal for advancing generative visual intelligence~\cite{20205longcat, yang2025longlive, chen2025skyreels}. Beyond short-clip synthesis, the objective is to simulate evolving visual worlds with persistent identity and temporal coherence-- powering applications in digital storytelling, robotics simulation, and world modeling. Despite rapid progress from large-scale diffusion–transformer architectures~\cite{Peebles2022DiT, SD3}, current systems are fundamentally limited by short temporal context and the quadratic cost of full attention, which constrain scalability and impede prompt-adaptive updates in real time~\cite{wan2025wan,kong2024hunyuanvideo,yang2024cogvideox}. 

% research progress on long video
Autoregressive (AR) diffusion offers an alternative-- it creates frames causally and enables streaming, interactive synthesis. However, pushing AR diffusion to minute- or hour-long horizons exposes persistent bottlenecks: error accumulation, motion drift, and content repetition~\cite{huang2025self,liu2025rolling,yang2025longlive}. Recent work addresses these issues by narrowing the train–test gap and enlarging the context. Self-Forcing~\cite{huang2025self} performs short AR rollouts during training, aligning the model with its own predictions via rolling key–value (KV) caches as short-term memory to suppress immediate drift. Rolling-Forcing~\cite{liu2025rolling} extends this with context rolling and cache reuse to propagate information over longer windows. LongLive~\cite{yang2025longlive} introduces a global attention-sink that reuses the earliest frames as fixed anchors, improving long-term stability. Yet these designs remain narrow in temporal adaptability: by repeatedly attending to static early frames, they freeze global memory, which encourages scene looping and repetition over long sequences. 

% memory as a motivation
Humans naturally rely on dynamic memory-- continuously updating past experience as new events unfold-- to reason coherently over long timescales. 
Inspired by this, recent world-model research~\cite{li2025vmem,huang2025voyager,yu2025context,xiao2025worldmem,po2025longcontextssm} augments video simulators with spatial memory. For example, VMem~\cite{li2025vmem} builds a surfel-indexed 3D view memory, while WorldMem~\cite{xiao2025worldmem} and Context-as-Memory~\cite{yu2025context} maintain cached scene tokens indexed by 3D camera poses for interactive generation. In contrast, long-video generation seeks open-ended, perceptually realistic synthesis over minutes, typically with free camera motion and highly dynamic scenes-- without an explicit world state. Consequently, such 3D-coupled memory designs transfer poorly to long-horizon, free-view settings. What’s missing is a continuously updated global memory in latent space-- one that jointly captures local motion and global scene evolution without explicit 3D assumptions. 

We introduce VideoSSM, an autoregressive long video diffusion model equipped with a hybrid memory architecture: a causal sliding-window local cache as the local memory and a State-Space Model (SSM) compressed global memory. Our design treats video generation as a recurrent dynamical process, where the compressed state via SSM continuously evolves with each generative step to retain and update the holistic scene state. To complement, the context window serves as local lossless memory, capturing temporal motion cues and visual details. 
This hybrid memory structure enables the model to balance stability and adaptability-- maintaining coherence across long horizons while dynamically responding to newly emerged content.  
Unlike the static attention-sink~\cite{yang2025longlive, liu2025rolling}, which fixes early-frame tokens as history anchors, VideoSSM updates memory continuously to preserve global context-- avoiding freezing global memory and content repetition.
Overall, it yields long-range coherence, content diversity, and interactive adaptability with linear-time cost scalability.

Extensive experiments on VBench~\cite{huang2024vbench} as short- and long-range benchmarks show that VideoSSM achieves state-of-the-art temporal consistency and motion stability among popular autoregressive video generators. At minute-scale horizons, it substantially reduces cumulative error, motion drift, and content repetition. Interactive evaluations with prompt switching further demonstrate smoother transitions, fewer residual semantics, and higher user preference via our user study. Together, these results position VideoSSM as a scalable, memory-aware framework for long-context and interactive video generation.

\section{Related Work}
\label{sec:related}

\subsection{Autoregressive Video Generation}
Different from methods adopting bidirectional attention to simultaneously generate all video frames~\cite{hacohen2024ltx, he2022latent, ho2022video, oshima2024ssm, yang2024cogvideox, blattmann2023stable, ha2018recurrent, mathieu2015deep}, AR video generation methods enable streaming and frame-wise video synthesis~\cite{huang2025self, 20205longcat,yin2025causvid}, making them particularly suitable for real-time and interactive prompt control~\cite{yang2025longlive, xiao2025worldmem}.
Similar to the AR paradigm in large language models, some methods first discretize videos into spatio-temporal tokens and then train next-token or next-block predictors~\cite{bruce2024genie,kondratyuk2024videopoet,ren2025next,wang2024loong,weissenborn2020scaling,yan2021videogpt,ge2022long,wu2022nuwa}. 
During inference,  spatio-temporal tokens are generated sequentially to compose a complete video. 
Other some works modify the diffusion objective by assigning different noise levels to individual frames during training~\cite{chen2024diffusionforcing, tulyakov2018mocogan}, where the current frame is noisier than previous ones. They thereby achieve AR-style generation by feeding synthesized frames back as context during inference~\cite{teng2025magi, yin2025causvid,gu2025long,chen2025skyreels}. 

\begin{figure*}[htbp!]
    \centering
    \includegraphics[width=0.95\linewidth]{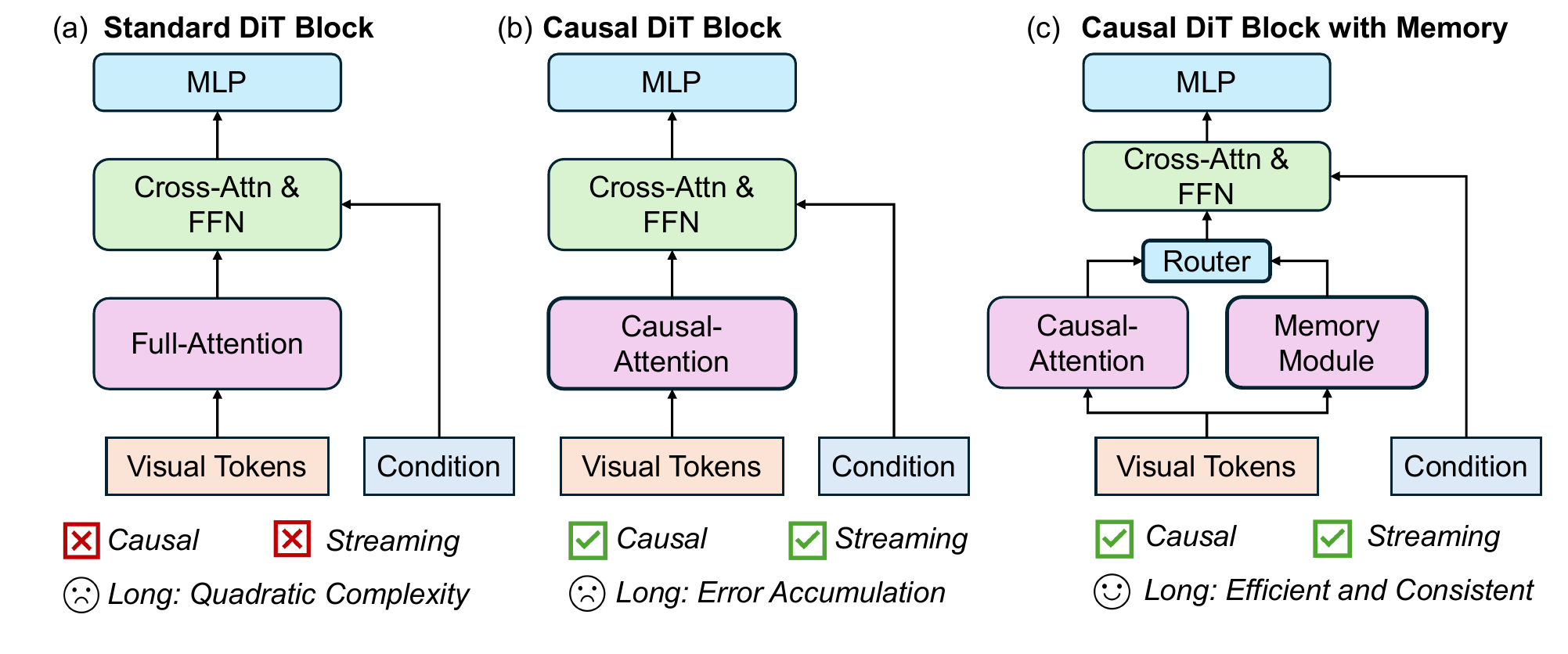}
    \vspace{-0.5cm}
    \caption{\textbf{Comparison of DiT block architectures for autoregressive video generation.} \textbf{(a)} Standard DiT block with full self-attention, which supports long-context modeling but lacks causality and streaming capability. \textbf{(b)} AR DiT block with masked causal attention, enabling autoregressive and streaming generation at the cost of weakened long-context consistency. \textbf{(c)} Our AR DiT block with a hybrid memory module and router, which combines local causal attention with a learnable global memory to achieve causal generation, streaming, and long-context support.
}
    \label{fig:dit_compare}
\vspace{-0.3cm}
\end{figure*}

\subsection{Long Video Generation} 

To extend diffusion models to longer durations, some works generate overlapping clips with temporal constraints or adopt hierarchical keyframe-to-interpolation pipelines~\cite{villegas2023phenaki, wu2022nuwa, wang2023lavie, chen2023seine, LCT_2025_ICCV}, which remains computation-heavy for long videos. Training-free extrapolation methods~\cite{zhao2025riflex, qiu2023freenoise, lufreelong,lu2025freelongpp} extend video length at inference by adjusting positional encodings, noise schedules, or temporal frequencies, yet unsuitable for real-time generation. AR models have shown strong scalability for variable-length and real-time generation~\cite{gu2025long, teng2025magi, chen2025sana, chen2024diffusionforcing, henschel2024streamingt2v, song2025history, chen2025skyreels, zhang2025packing, oshima2024ssm}. To mitigate long-term drift, AR-diffusion methods adopt train–test alignment via rollout with KV caching~\cite{huang2025self} and DMD loss~\cite{yin2024onestep}, extend context through KV rolling or recaching~\cite{cui2025self, yang2025longlive}, and integrate attention with sink tokens~\cite{liu2025rolling}. Yet when extend to minute-scale, these methods still suffer from severe content repetition.

\subsection{Memory Mechanism in Generative Models}
Memory in generative models can be categorized into local and global. Local memory is typically implemented via sliding-window attention and a KV cache~\cite{cheng2025playing,zhai2025stargen,gu2025long} but inherently loses information outside the window~\cite{zhang2025packing}. Global memory aggregates history outside the local window. Some methods utilize early-frame for attention sinks to provide a persistent reference~\cite{xiao2023streamingllm, yang2025longlive, liu2025rolling, xu2025streamingvlmrealtimeunderstandinginfinite}, but they tend to freeze the global state and cause content repetition. Other approaches select key historical tokens from a memory pool~\cite{cai2025moc} or compress the history into a reusable implicit state to maintain long-range context~\cite{oshima2024ssm, fang2025artificial,po2025longcontextssm}. Specifically, geometric memory, used in recent world models, solves long-term consistency with explicit 3D structures and precise camera control~\cite{yu2025context, wu2025video,huang2025voyager,li2025vmem}. While showing potential for interactive simulation with viewpoint revisits, the 3D-coupled memory transfers poorly to open-ended, long-horizon, and free-view synthesis.
In this work, we introduce a SSM that functions as a dynamic, continuously evolving global memory.

\begin{figure*}
    \centering
    \includegraphics[width=0.9\linewidth]{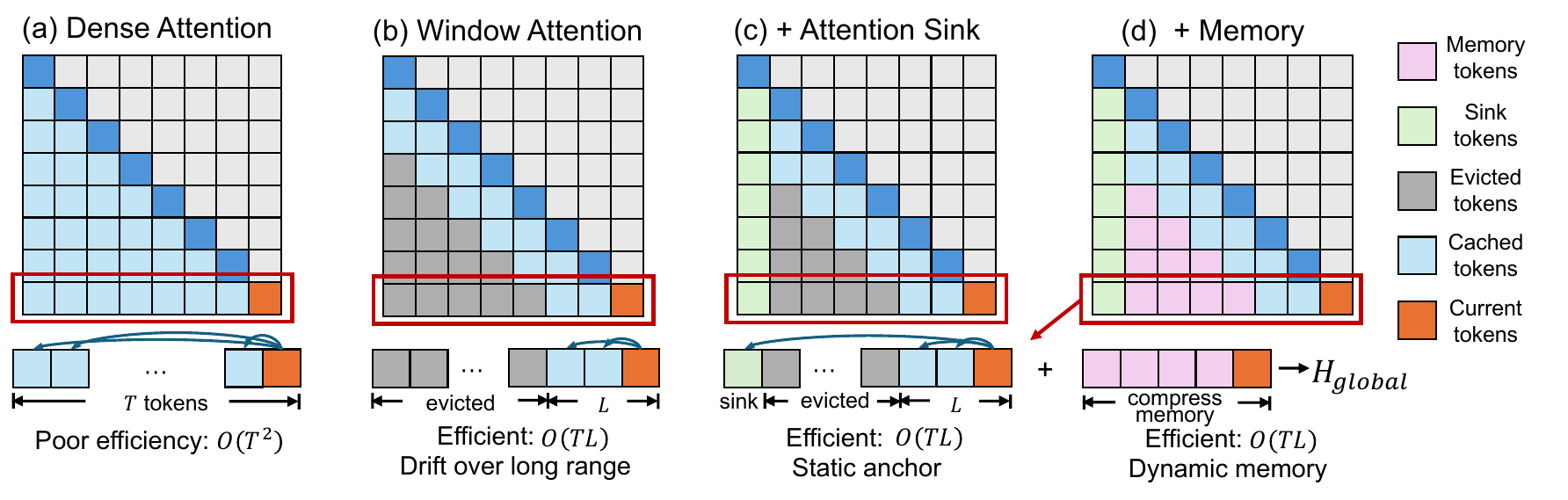}
    % \vspace{-5mm}
    \caption{\textbf{Illustration of attention mechanisms in AR DiT.} 
    Let $T$ be the video token length and $L$ the sliding-window size. 
    \textbf{(a) Causal Attention}: Each query attends to all past tokens. It captures the full context with quadratic O(T²) complexity, impractical for long sequences.
    \textbf{(b) Window Attention:} Localized attention within a local sliding window. It enables efficient O(TL) complexity for streaming but causes information drift as early tokens are evicted.
    \textbf{(c) Attention Sink:} Adds fixed initial ``sink" tokens to the window. It improves long-range consistency with O(TL) complexity, but the static memory leads to repetitive generation and fails to adapt to new content
    \textbf{(d) Ours (Hybrid Memory):}  Augments window attention with a learnable memory that compresses evicted tokens. This maintains O(TL) efficiency while providing a dynamic global context, balancing long-term consistency and adaptability.
}
    \label{fig:attn_compare}
    \vspace{-0.2cm}
\end{figure*}

\section{Preliminary: From DiT to AR DiT}
\label{sec:prelim}
\paragraph{Diffusion Transformers for Video.}
We begin by reviewing the standard Diffusion Transformer (DiT) formulation for video generation.  
Given a clean video sample $\mathbf{x}_0 \sim q(\mathbf{x}_0)$, the forward diffusion process gradually corrupts $\mathbf{x}_0$ via a Markov chain:
\begin{equation}
q(\mathbf{x}_t \mid \mathbf{x}_{t-1}) 
= \mathcal{N}\!\left(
\mathbf{x}_t;\, 
\sqrt{1 - \beta_t}\,\mathbf{x}_{t-1},\,
\beta_t \mathbf{I}
\right),
\end{equation}
where $\beta_t$ denotes the noise schedule. 
The reverse process learns to approximate the true posterior using a denoiser parameterized by a noise-prediction network \( \boldsymbol{\epsilon}_\theta \), trained with the standard objective:
\begin{equation}
\mathcal{L}
= \mathbb{E}_{\mathbf{x}_0, \boldsymbol{\epsilon}, t}
\left\|
\boldsymbol{\epsilon} - \boldsymbol{\epsilon}_\theta(\mathbf{x}_t, t)
\right\|^2, \
\mathbf{x}_t 
= \sqrt{\bar{\alpha}_t}\,\mathbf{x}_0 
  + \sqrt{1 - \bar{\alpha}_t}\,\boldsymbol{\epsilon},
\end{equation}
with $\bar{\alpha}_t = \prod_{s=1}^t (1 - \beta_s)$.  
In standard DiTs (Fig.~\ref{fig:dit_compare}~(a)), $\boldsymbol{\epsilon}_\theta$ is implemented by a bidirectional vision Transformer that operates on spatiotemporal tokens obtained from video frames, and each token attends to all others through full self-attention.

\paragraph{Autoregressive DiT for Long Video.}
To enable long-horizon video generation, DiT is converted into a causal autoregressive AR model (Fig.~\ref{fig:dit_compare}~(b)). 
Let $\mathbf{c}_t$ denote a conditioning signal derived from past frames. The denoiser then becomes conditional, $\boldsymbol{\epsilon}_\theta(\mathbf{x}_t, t, \mathbf{c}_t)$, and the Transformer backbone is upgraded to a \emph{causal AR DiT} by restricting self-attention along the temporal sequence, so that the new frame can only attend to previous frame tokens. 
At inference time, the model generates video in an autoregressive manner: each newly synthesized frame is fed back to update $\mathbf{c}_t$, so that future frames are conditioned only on previously generated content rather than ground-truth frames. 

However, pushing AR diffusion to minute- or hour-long horizons leads to error accumulation, motion drift, and content repetition.
In this work, we augment the AR DiT in Fig.~\ref{fig:dit_compare}~(b) with a hybrid memory architecture (Fig.~\ref{fig:dit_compare}~(c)).

\section{VideoSSM}
\label{sec:method}

We propose VideoSSM, a state-space long video model that augments autoregressive DiT with a hybrid memory architecture for coherent and scalable long-horizon synthesis.
Section~\ref{sec:constraints} analyzes the limitations of existing AR architectures by examining their attention and caching strategies (Fig.~\ref{fig:attn_compare}).
Section~\ref{sec:hybrid_memory} introduces our hybrid memory design, which combines a learnable global memory with local sliding-window context.
Section~\ref{sec:training} presents a causal distillation framework that efficiently transfers knowledge from a bidirectional teacher to our causal AR model. 

\subsection{Motivations}
\label{sec:constraints}

In an AR DiT model, causal attention enables temporally streaming generation but requires maintaining a KV cache whose size increases linearly with the number of past tokens (Fig.~\ref{fig:attn_compare}(a)).
To control memory growth, many systems adopt sliding-window attention with a rolling KV cache~\cite{xiao2023streamingllm,xu2025streamingvlmrealtimeunderstandinginfinite} (Fig.~\ref{fig:attn_compare}(b)).
This design is memory- and latency-efficient but suffers from drifting and error accumulation when early tokens are evicted~\cite{xiao2023streamingllm}.
To stabilize long-range dependencies, attention-sink mechanisms preserve a small set of initial ``sink'' tokens (Fig.~\ref{fig:attn_compare}(c)).
In long-video generation, the earliest frames are often used as sink tokens~\cite{yang2025longlive,liu2025rolling}, combined with recent tokens to ensure stable attention computation.

However, our experiments show that sink-based attention frequently produces \textbf{repetitive content or frozen generation patterns}, especially in long videos.
The fixed sink tokens over-stabilize the KV cache, overshadowing the contributions of the evolving sliding-window context.

\subsection{Hybrid State-Space Memory}
\label{sec:hybrid_memory}

We propose VideoSSM, an autoregressive diffusion model equipped with a hybrid memory architecture (Fig.~\ref{fig:attn_compare}(d)). Rather than discarding out-of-window tokens or relying on fixed attention sinks, we introduce a dedicated memory module that explicitly manages both short-term and long-term information.

Inspired by the hierarchical structure of human memory~\cite{atkinson1968human,baddeley2020working,cowan2008differences}-- where working memory retains fine details and long-term memory stores compressed, abstract representations, we decompose model memory into two complementary components and integrate them into the attention mechanism: 
\begin{itemize}
\item \textbf{Local Memory}. A causal attention window with cached KV states (Fig.~\ref{fig:attn_compare}(b)) that preserves precise, lossless representations, essential for capturing fine-grained motion and appearance details (Sec.~\ref{sec:local}).
\item \textbf{Global Memory}. An evolving memory module (Fig.~\ref{fig:dit_compare}(c)) that absorbs tokens evicted from the local window and recurrently compresses them into a compact, fixed-size state (Fig.~\ref{fig:attn_compare}(d)), providing a continuously updated summary of all past context (Sec.~\ref{sec:global}).
\end{itemize}

As illustrated in Fig.~\ref{fig:attn_compare}(d), this hybrid attention mechanism allows the model to access the entire history while maintaining $O(TL)$ complexity and full streaming capability. The design meets the core requirements of modern long-video generators: it is causal, naturally streamable, and capable of leveraging long-range temporal context efficiently. 

\subsubsection{Local Memory: Sliding Window Self-Attention}
\label{sec:local}
% The local memory component functions as a standard causal self-attention mechanism with a fixed-size sliding window, as is shown in Fig.~\ref{fig:local_memory}.
Let $L$ be the sliding-window size.
Given the input hidden state $\mathbf H^{\text{in}}_t$ for the current frame $t$, queries ($\mathbf Q_t$), keys ($\mathbf K_t$), and values ($\mathbf V_t$) are computed as:

\begin{equation}
\{\mathbf Q_t, \mathbf K_t, \mathbf V_t \} =\{   \mathbf H^{\text{in}}_t \mathbf W_Q,  \mathbf H^{\text{in}}_t \mathbf W_K, \mathbf H^{\text{in}}_t \mathbf W_V\},
\end{equation}
where $\mathbf K_t$ and $\mathbf V_t$ pairs are appended to the local KV Cache. The cache retains only key-value pairs of the sinking token and the $L$ most recent tokens, forming $\mathbf K^{\text{local}}_t = [\mathbf K_{\text{sink}}, \mathbf K_{t-L+1}: \mathbf K_{t}]$ and $\mathbf V^{\text{local}}_t= [\mathbf V_{\text{sink}}, \mathbf V_{t-L+1}:\mathbf V_{t}]$.  $\mathbf H^{\text{local}}_t$ based on local memory is computed with a standard causal self-attention mechanism:

\begin{equation}
\mathbf H^{\text{local}}_t = \text{SelfAttention}(\mathbf Q_t, \mathbf K^{\text{local}}_t, \mathbf V^{\text{local}}_t). 
\end{equation}

\begin{figure}[t]
    \centering
    \includegraphics[width=1\linewidth]{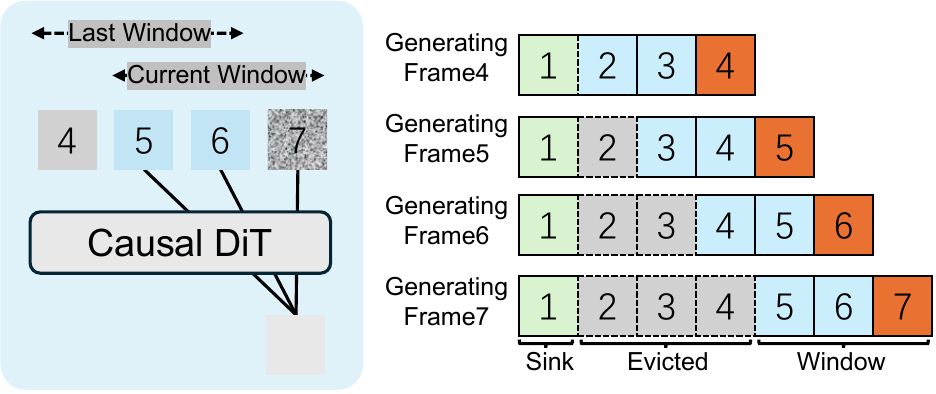}
    \caption{Illustration of how sink, evicted, and window tokens are arranged at different timesteps in a causal DiT with sliding-window attention. Here window length $L=3$.}
    \label{fig:local_memory}
    \vspace{-5mm}
\end{figure}

\subsubsection{Global Memory: Dynamic State Computation}
\label{sec:global}
The global memory module compresses the full history of out-of-window (evicted) tokens into a fixed-size, continuously evolving representation. It operates through four key components: gate caching, state updates, memory retrieval, and output gating.  

\vspace{0.1in}\noindent \textbf{Synchronized Gate Caching.}
To integrate information from an evicted token into the global memory, we maintain a recurrent compressed state governed by two learnable gates at each timestep~$t$: an \emph{injection gate} $\boldsymbol{\beta}_t$ and a \emph{decay gate} $\boldsymbol{\alpha}_t$. $\boldsymbol{\alpha}_t, \boldsymbol{\beta}_t \in \mathbb{R}^d$, both matching the dimensionality $d$ of latent tokens.
The injection gate controls how strongly the evicted token should update the global state, while the decay gate determines how quickly past memory should fade.
Both gates are computed from the hidden state $\mathbf{H}^{\text{in}}_t$ before the token exits the local window: 

\begin{figure}[!t]
    \centering
    \includegraphics[width=0.95\linewidth]{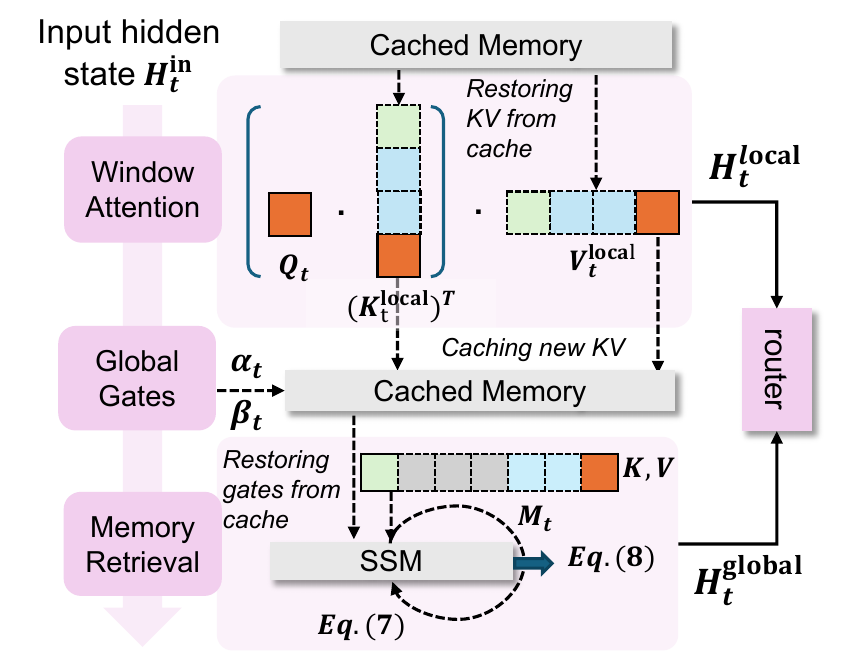}
    % \vspace{-5mm}
    \caption{Architecture of the proposed hybrid memory module. The input $H^{\text{in}}_t$ is processed in two streams. The \textbf{local path} (top) uses windowed attention with a sliding KV cache to compute $H^{\text{local}}_t$. The \textbf{global path} (bottom) uses a State-Space Model (SSM) to recurrently compress historical information into a memory state $M$, which is retrieved to produce $H^{\text{global}}_t$. A \textbf{router} then dynamically fuses the local and global outputs.}
    \label{fig:global_memory}
    % \vspace{-2mm}
\end{figure}

\begin{align}
\bm \beta_t &= \sigma(\mathbf W_\beta \mathbf H^{\text{in}}_ t) \label{eq:beta_gate} \\
\bm \alpha_t &= -\exp(\mathbf A) \cdot \text{SoftPlus}(\mathbf W_\alpha \mathbf H^{\text{in}}_ t + \mathbf{B}) \label{eq:alpha_gate}
\end{align}
where $\sigma$ is the sigmoid function and \text{SoftPlus} is a smooth variant of the ReLU activation function; $\mathbf{A}$, $\mathbf{W}_\alpha$, and $\mathbf{W}_\beta$ are learnable weights; and $\mathbf{B}$ is learnable bias.
The gates ${\boldsymbol{\alpha}_t, \boldsymbol{\beta}_t}$ are stored in a Gates Cache that is updated in sync with the rolling KV cache.

\vspace{0.1in}\noindent\textbf{Global Memory State Update.} 
For a query at time~$t$, tokens outside the $L$-length window, indexed by $[\text{sink}+1 : t-L]$ are considered evicted. 
We denoted $ \{\mathbf K,\mathbf V,\bm \alpha, \bm \beta \}^{\text{evt}}_t = \text{avg}[\mathbf \{\mathbf K,\mathbf V,\bm \alpha, \bm \beta \}_{\text{sink}+1} : \mathbf \{\mathbf K,\mathbf V,\bm \alpha, \bm \beta \}_{t-L}]$. 
A summary of all information up to the most recently evicted token is compacted into the global state $\mathbf{M}_t$.
We update $\mathbf{M}_t$ using the Gated $\Delta$-rule~\cite{yang2412gated}, which extracts only the \emph{novel} component of the incoming information before integrating it into the state: 

\begin{equation}
\begin{aligned}
\mathbf V^{\text{evt}}_{\text{new},t} &= \mathbf V^{\text{evt}}_t - \text{Predict}(\mathbf M_{t-1}, \mathbf K^{\text{evt}}_t, \bm \beta^{\text{evt}}_t), \\
\mathbf{M}_t &= \exp(\bar{\mathbf{g}}_t) \cdot \mathbf{M}_{t-1} + \mathbf{K}^{\text{evt}}_t \cdot ({\mathbf{V}^{\text{evt}}_{\text{new},t}})^T,
\label{eq:state_update}
\end{aligned}
\end{equation}
  
where $\mathbf{M}_0 = \mathbf{0}$, $\text{Predict}(\cdot)$ estimates the predictable portion of the evicted value from the previous state $\mathbf{M}_{t-1}$, so that $\mathbf{V}^{\text{evt}}_{\text{new},t}$ retains only the unpredictable, novel component of the input, and $\bar{\mathbf{g}}_t = \sum_{s=0}^{t} \boldsymbol{\alpha}^{\text{evt}}_s$ is a cumulative negative gate (with $\boldsymbol{\alpha}^{\text{evt}}_s < 0$) that controls state decay.
The first term, $\exp(\bar{\mathbf{g}}_t)\cdot \mathbf{M}_{t-1}$, ensures controlled forgetting and long-term stability, while the second term integrates only the unpredictable, new information.
By combining selective update and gated decay, the global memory maintains a compact, stable, and continuously evolving representation that grows with the video rather than collapsing or drifting-- providing robust long-range context for AR diffusion.

\vspace{0.1in}\noindent\textbf{Global Memory Retrieval.}
We retrieve a query-aligned memory response by projecting the current query $\mathbf{Q}_t$ onto the compressed state $\mathbf{M}_{t}$, which is then refined by an output gate that regulates how much global information is exposed to the backbone. The final global context $\mathbf{H}^{\text{global}}_t$ is computed as:
\begin{equation}
\begin{aligned}
% \mathbf{H}^{\text{memory}}_t &= \mathbf{Q}_t^T\mathbf{M}_{t}, \\
\mathbf{g}^{\text{out}}_t &= \text{Linear}(\mathbf{H}^{\text{in}}_t), \\
\mathbf{H}^{\text{global}}_t &= \text{Swish}(\mathbf{g}^{\text{out}}_t \odot \text{RMSNorm}(\mathbf{Q}_t\mathbf{M}_{t})),
\end{aligned}
\label{eq:output_gate}
\end{equation}
where $\mathbf{H}^{\text{in}}_t$ is the current hidden state, $\mathbf{g}^{\text{out}}_t$ is a vector of output gate coefficients controlling information flow, and $\text{Swish}(x) = x \cdot \sigma(x)$ denotes the Swish activation.
It first normalizes the memory response with RMSNorm, then applies element-wise gating, allowing the model to selectively filter and modulate global information before passing it to subsequent layers.

\subsubsection{Position-Aware Gated Fusion} 
\label{sec:router}

Finally, we fuse the local and global memory streams. We introduce a \emph{position-aware router} that dynamically controls the strength of global memory injection. This is crucial for stable generation: early in the sequence, the model should rely primarily on local information, whereas global memory should play a larger role as more context accumulates.

We define a relative position ratio $\rho_t = (t+1)/T$, where $t$ is the current frame index and $T$ is the total context length. A memory gate $\boldsymbol{\gamma}_t \in \mathbb{R}^d$, matching the token dimensionality $d$, is then computed as 
\begin{equation}
\bm \gamma_t = \sigma\!\left( \boldsymbol{w}_{\text{router}} \log(\rho_t) + \boldsymbol{b}_{\text{router}} \right), 
\end{equation}
where $\boldsymbol{w}_{\text{router}}, \boldsymbol{b}_{\text{router}} \in \mathbb{R}^d$ are learnable vectors.
As $t \to 0$, we have $\rho_t \to 0$, $\log(\rho_t) \to -\infty$, and thus $\boldsymbol{\gamma}_t \to \mathbf{0}$, effectively suppressing the global memory at the beginning of the sequence.
As $t \to T$, $\rho_t \to 1$, $\log(\rho_t) \to 0$, and $\boldsymbol{\gamma}_t$ approaches the learned value $\sigma(\boldsymbol{b}_{\text{router}})$,  enabling the global memory stream.

The final fused hidden state $\mathbf{H}^{\text{fused}}_t$ is a simple gated sum:
\begin{equation}
\mathbf{H}^{\text{fused}}_t = \mathbf{H}^{\text{local}}_t + \gamma_t \cdot \mathbf{H}^{\text{global}}_t, 
\end{equation}
which is then passed to the subsequent cross-attention and feed-forward layers of the DiT block.
This fusion mechanism allows VideoSSM to retain the $O(L)$ efficiency of sliding-window attention while incorporating a dynamic, compressed global context—mitigating both the catastrophic forgetting of pure windowed attention and the rigidity of static attention sinks, and yielding dynamics that are both temporally coherent over long horizons and responsive to evolving scene content. 

\subsection{Training with Memory}
\label{sec:training}

To enable real-time generation, we distill a high-fidelity teacher into a causal framework based on Self-Forcing \cite{huang2025self}. Our model incorporates a hybrid memory for streaming consistency and uses a rolling memory recipe for efficient long-horizon training and interactive prompt switching.

\vspace{0.1in}\noindent\textbf{Stage 1: Causal Model Distillation.}
We initialize the causal student model $G_{\theta}$ from a pre-trained bidirectional teacher $T_{\phi}$ (\textit{i.e.}, Wan 2.1~\cite{wan2025wan}) following the CausVid strategy~\cite{yin2025causvid}. 
Using the teacher’s ODE sampling trajectories, the student is trained to match the teacher’s short-clip expertise on 5-second segments. 
The student regresses these trajectories causally, minimizing $\mathcal{L} = \|\hat{\mathbf{x}}_0 - T_{\phi}(\mathbf{x}_t, t)\|^2$, where $\hat{\mathbf{x}}_0 = G_{\theta}(\mathbf{x}_t, t)$. Gradients are computed selectively and propagated across steps to the hybrid memory, mitigating exposure bias via self-generated histories.
This stage equips $G_{\theta}$ with high-quality short-term dynamics before introducing long-horizon autoregressive behavior. By incorporating our hybrid memory, the model gains long-range capabilities even trained only on short segments. 

\vspace{0.1in}\noindent\textbf{Stage 2: Long Video Training.}
The second stage mitigates long-horizon degradation by training the hybrid memory for effective operation in streaming, autoregressive scenarios.
We extend the SF-style distillation using DMD, allowing the student model to self-correct errors during training that mimics inference with rolling KV caches and memory.

\begin{enumerate}
    \item \textbf{Long Self-Rollout:}
    $G_{\theta}$ autoregressively produces a long sequence $\hat{x}^{1:N}$ (e.g., $N=60$ seconds) in chunks, surpassing the teacher's 5-second horizon. It fills its local \texttt{KV Cache} and global \texttt{Memory Cache} (via $\beta$ and $\alpha$ gates) exclusively from self-generated outputs, employing a rolling KV cache with global memory to sustain fixed context length and avert drift, thereby emulating inference.
    
    \item \textbf{Windowed Teacher Correction:} Uniformly sample a short window $K$ ($K=5$ seconds). Apply the DMD loss over this window:
    \begin{equation}
    \label{eq:dmd}
    \mathcal{L}_{\text{DMD}} = \mathbb{E}_{t, i \sim \text{Unif}(1, N-K)} \left[ \nabla_{\theta} KL(p_{\theta,t}^{S}(z_i) || p_{t}^{T}(z_i)) \right]
    \end{equation}
    where $z_i$ is the window starting at frame $i$. This harnesses the teacher's short-clip proficiency to rectify long-range errors, facilitating recovery from degraded states.
\end{enumerate}

\section{Experiments}
\label{sec:experiments}

\subsection{Settings}
We implement VideoSSM based on the Wan 2.1-T2V-1.3B model~\cite{wan2025wan}, a flow-matching model that generates 5-second videos at 16 FPS with 832$\times$480 resolution. The distilled chunk-wise autoregressive model uses 4-step diffusion and with chunk size of 3, generating a chunk of 3 latent frames at a time.
For both ODE initialization and Long Video Distillation, we sample text prompts from a filtered and LLM-extended version of VidProM~\cite{wang2024vidprom}.

We evaluate on VBench~\cite{huang2024vbench} for short videos (5 seconds) and long videos (minute-long), measuring dimensions including subject consistency, background consistency, motion smoothness, aesthetic quality, and imaging quality. 
In the user study, we collected preferences from 40 participants. Each participant was shown 8 prompts, and for each prompt, they ranked 4 generated one-minute videos generated with three AR video generation models: Self-forcing~\cite{huang2025self}, CausVid~\cite{yin2025causvid}, Longlive~\cite{yang2025longlive}, and ours.

\subsection{Video Quality Evaluation}
We first evaluate short-video (5-second) generation on the VBench benchmark~\cite{huang2025self}. As shown in Table~\ref{tab:short_vbench}, we compare VideoSSM against leading AR models, including few-step distilled generators~\cite{yin2025causvid, huang2025self} and their long-range variants~\cite{yang2025longlive, cui2025self, liu2025rolling}, alongside other strong baselines~\cite{deng2024autoregressive, jin2024pyramidal, chen2025skyreels, teng2025magi}. For reference, we also list SOTA bidirectional models~\cite{hacohen2024ltx, wan2025wan}. 
VideoSSM achieves the highest Total (83.95) and Quality (84.88) scores among all AR models, outperforming strong competitors like LongLive and the 4.5B-parameter MAGI-1. This demonstrates that our hybrid memory mechanism effectively enhances short-video fidelity.

\begin{figure*}
    \centering
    \includegraphics[width=1\linewidth]{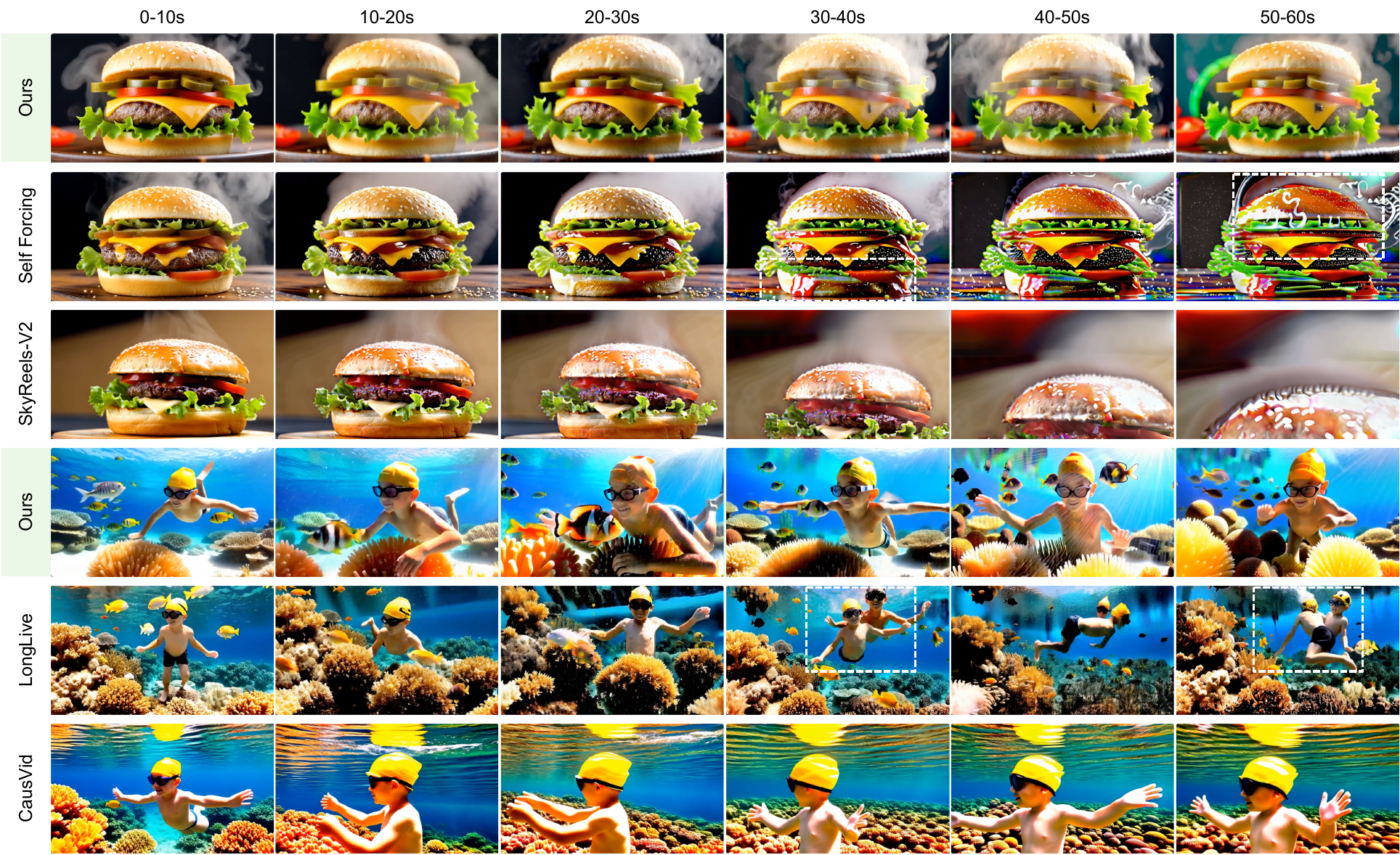}
    \caption{Qualitative comparison of 60s-long video generation. Baseline methods with windowed attention often suffer from error accumulation, leading to drifting artifacts, while methods using attention sink may produce repeated content or nearly static scenes. Our approach generates videos with more coherent motion and stable temporal consistency.}
    \label{fig:quality}
    \vspace{-4mm}
\end{figure*}

\begin{table}[thbp]
\centering
\caption{Comparison with relevant baselines. We compare VideoSSM with representative open-source video generation models of similar parameter sizes and resolutions.\protect\footnotemark}
% \vspace{-2mm}
\resizebox{\linewidth}{!}{
\begin{tabular}{lrrrr}
\toprule
Model & \#Params & \multicolumn{3}{c}{Evaluation scores $\uparrow$} \\
\cmidrule{3-5}
& & Total & Quality & Semantic \\
\midrule
\rowcolor{lightgray}
\multicolumn{5}{l}{\textit{Bidirectional Diffusion models}} \\
\textcolor[gray]{0.5}{LTX-Video~\cite{hacohen2024ltx}} & \textcolor[gray]{0.5}{1.9B} & \textcolor[gray]{0.5}{80.00} & \textcolor[gray]{0.5}{82.30} & \textcolor[gray]{0.5}{70.79} \\
\textcolor[gray]{0.5}{Wan2.1~\cite{wan2025wan}} & \textcolor[gray]{0.5}{1.3B} & \textcolor[gray]{0.5}{84.26} & \textcolor[gray]{0.5}{85.30} & \textcolor[gray]{0.5}{80.09} \\
\midrule
\rowcolor{lightgray}
\multicolumn{5}{l}{\textit{Autoregressive models}} \\
SkyReels-V2~\cite{chen2025skyreels} & 1.3B & 82.67 & 84.70 & 74.53 \\
MAGI-1~\cite{teng2025magi} & 4.5B & 79.18 & 82.04 & 67.74 \\
CausVid~\cite{yin2025causvid} & 1.3B & 81.20 & 84.05 & 69.80 \\
NOVA~\cite{deng2024autoregressive} & 0.6B & 80.12 & 80.39 & 79.05 \\
Pyramid Flow~\cite{jin2024pyramidal} & 2B & 81.72 & 84.74 & 69.62 \\
Self Forcing~\cite{huang2025self} & 1.3B & 83.00 & 83.71 & 80.14 \\
% Self Forcing, frame-wise (Huang et al., 2025) & 1.3B & 84.26 & 85.25 & 80.30 \\
LongLive~\cite{yang2025longlive} & 1.3B & 83.52 & 84.26 & 80.53 \\
Self Forcing ++~\cite{cui2025self} & 1.3B & 83.11 & 83.79 & 80.37 \\
Rolling Forcing~\cite{liu2025rolling} & 1.3B & 81.22 & 84.08 & 69.78 \\
\midrule
VideoSSM (Ours) & 1.4B & \textbf{83.95} & \textbf{84.88} & 80.22 \\
\bottomrule
\end{tabular}
}
\vspace{-3mm}
\label{tab:short_vbench}
\end{table}

\footnotetext{LongLive results are based on our implementation; other results are taken from \cite{huang2025self, cui2025self, liu2025rolling}.}

\subsection{Long Video Generation}
To assess long-video capabilities, we evaluate minute-long generations using single prompts on VBench. 
We evaluate the train-short test-long setting, where models trained on 5-second clips must generalize to long videos, a regime that exposes catastrophic drift. As shown in Table~\ref{tab:ar_performance_cyan}, VideoSSM achieves the highest Subject and Background Consistency among all AR models, demonstrating the effectiveness of our hybrid memory in preventing error accumulation.
Importantly, this consistency does not collapse into static or frozen outputs, as VideoSSM attains a markedly higher Dynamic Degree (50.50) than LongLive and Self Forcing, demonstrating its ability to maintain long-term coherence while preserving natural temporal evolution.

Figure~\ref{fig:quality} provides qualitative validation. In the \textit{burger} example, VideoSSM (Ours) maintains the subject's identity and structure for the full 60 seconds, whereas SkyReels-V2 suffers complete content collapse and Self Forcing exhibits severe drifting. The \textit{underwater} scene is more revealing: our model successfully captures the dynamic, forward-swimming motion while maintaining high subject consistency. In contrast, competing methods fail this balance. CausVid avoids drift but succumbs to motion stagnation, with the child becoming nearly static in later frames. LongLive, which uses a fixed attention-sink, initially preserves the subject but suffers degradation where it hallucinates a second instance of the boy. This combined evidence proves our hybrid memory's superior ability to enable stable, high-fidelity, and truly dynamic generation far beyond the short clips seen during training.

\begin{table}[t]
\centering
\caption{Performance comparisons of AR models on 60s long videos. Bold highlights the highest, underline the second highest.}
% \vspace{-2mm}
% \small
\resizebox{\linewidth}{!}{
\renewcommand{\arraystretch}{1.2}
\begin{tabular}{lccc}
\toprule
Metric &
\makecell{Self \\ Forcing} &
\makecell{LongLive} &
\makecell{\textbf{VideoSSM }\\ \textbf{(Ours)}} \\
\midrule
\textit{Temporal Flickering} $\uparrow$ &
\cellcolor{cyan!60}\textbf{97.86} & 
\cellcolor{cyan!20}97.24 & 
\cellcolor{cyan!40}\underline{97.70} \\

\textit{Subject Consistency} $\uparrow$ &
\cellcolor{cyan!20}88.25 & 
\cellcolor{cyan!40}\underline{91.09} & 
\cellcolor{cyan!60}\textbf{92.51} \\

\textit{Background Consistency} $\uparrow$ &
\cellcolor{cyan!20}91.73 & 
\cellcolor{cyan!40}\underline{93.23} & 
\cellcolor{cyan!60}\textbf{93.95} \\

\textit{Motion Smoothness} $\uparrow$ &
\cellcolor{cyan!60}\textbf{98.67} & 
\cellcolor{cyan!20}98.38 & 
\cellcolor{cyan!40}\underline{98.60} \\

\textit{Dynamic Degree} $\uparrow$ &
\cellcolor{cyan!20}35.00 & 
\cellcolor{cyan!40}\underline{37.50} & 
\cellcolor{cyan!60}\textbf{50.50} \\

\textit{Aesthetic Quality} $\uparrow$ &
\cellcolor{cyan!40}\underline{60.02} & 
\cellcolor{cyan!20}55.74 & 
\cellcolor{cyan!60}\textbf{60.45} \\

\bottomrule
\end{tabular}
}

\label{tab:ar_performance_cyan}
\vspace{-3mm}
\end{table}

\subsection{Interactive Video Generation}
\begin{figure}
    \centering
    \includegraphics[width=1\linewidth]{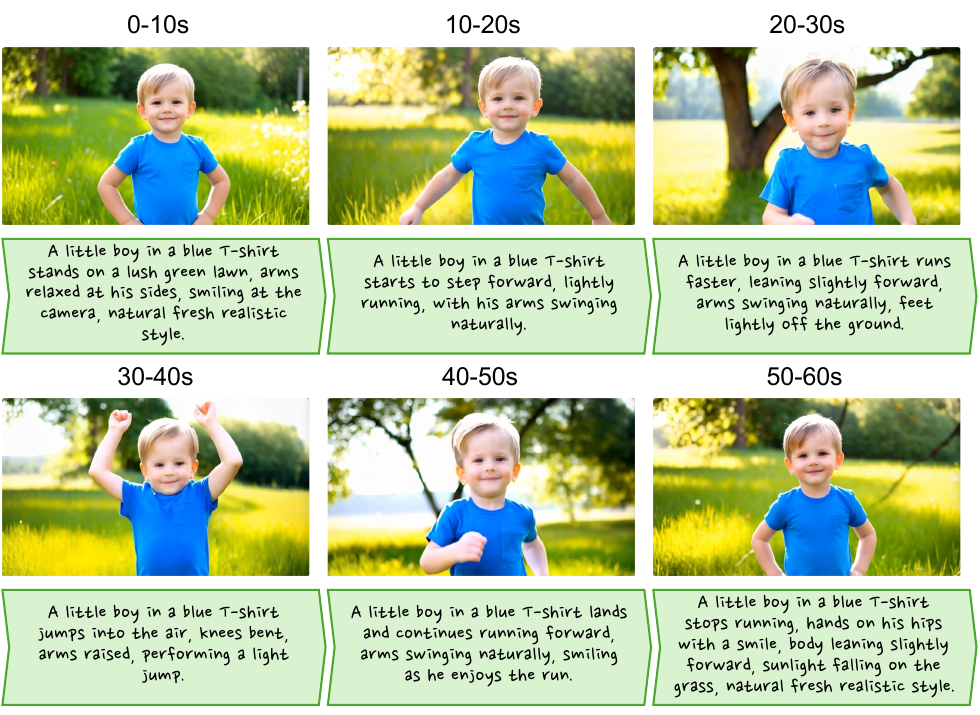}
    \vspace{-5mm}
    \caption{Demonstration of interactive long video generation.}
    \label{fig:interactive}
    \vspace{-4mm}
\end{figure}

Our model also supports interactive long video generation with seamless prompt switching after long-horizon training. By enabling KV recache\cite{yang2025longlive}, the system can efficiently refresh its internal local memory when the user provides new instructions, preventing outdated semantics from lingering and ensuring that each transition responds cleanly to the updated prompt. 
This allows the model to maintain scene coherence while adapting to new narrative directives in real time.
As illustrated in Fig.~\ref{fig:interactive}, VideoSSM produces smooth, natural, and dynamically coherent transitions across prompt changes. 
Additional qualitative examples are provided in the supplementary material.

\subsection{User Study}
To evaluate perceptual qualities beyond automated metrics, we conducted a user study with 40 participants. We generated 32 unique, minute-long videos by running our method and three baselines (LongLive~\cite{yang2025longlive}, Self Forcing~\cite{huang2025self}, CausVid~\cite{yin2025causvid}) on 8 different text prompts. For each prompt, participants were shown the four resulting videos in a randomized order and asked to rank them from 1 (best) to 4 (worst). The ranking criteria included overall visual quality, temporal and physical consistency, and adherence to the prompt. As shown in Table~\ref{tab:vote_percentages_avg}, our approach achieved the highest preference. VideoSSM received the most Rank 1 votes (41.07\%) and secured the best (lowest) average rank of 1.85.
While LongLive also achieves high consistency via a static attention-sink mechanism (low Dynamic Degree 37.50) that may minimize perceptual artifacts, VideoSSM’s higher Dynamic Degree enables complex, non-repetitive motion, yielding superior user preference by balancing dynamic realism with long-term consistency.

\begin{table}[t]
\centering
\small
\caption{Vote percentages for each model across different ranks. Cell color intensity indicates higher percentages. The last column shows the average rank.}
\vspace{-3mm}
\resizebox{\linewidth}{!}{
\renewcommand{\arraystretch}{1.2}
\begin{tabular}{lccccc}
\toprule
\textbf{Model} & \textbf{Rank 1 (\%)} & \textbf{Rank 2 (\%)} & \textbf{Rank 3 (\%)} & \textbf{Rank 4 (\%)} & \textbf{Avg Rank} \\
\midrule
Self Forcing & 
\cellcolor{cyan!15}11.79 & 
\cellcolor{cyan!15}13.21 & 
\cellcolor{cyan!25}23.21 & 
\cellcolor{cyan!55}51.79 & \cellcolor{green!5}3.18 \\

CausVid & 
\cellcolor{cyan!10}7.50 & 
\cellcolor{cyan!20}16.07 & 
\cellcolor{cyan!40}42.14 & 
\cellcolor{cyan!35}34.29 & \cellcolor{green!10}3.03 \\

LongLive & 
\cellcolor{cyan!50}39.64 & 
\cellcolor{cyan!45}36.43 & 
\cellcolor{cyan!15}15.00 & 
\cellcolor{cyan!5}8.93 & \cellcolor{green!30}1.92 \\

\textbf{Ours} & 
\cellcolor{cyan!55}\textbf{41.07} & 
\cellcolor{cyan!40}\textbf{34.29} & 
\cellcolor{cyan!20}19.64 & 
\cellcolor{cyan!5}\textbf{5.00} & \cellcolor{green!40}1.85 \\
\bottomrule
\end{tabular}
}

\label{tab:vote_percentages_avg}
\vspace{-3mm}
\end{table}

\section{Conclusion}
\label{sec:discussion}
In this work, we introduced VideoSSM, a long-video autoregressive diffusion model that reframes autoregressive diffusion as a recurrent dynamical process with hybrid memory: an SSM as evolving global memory and a context window as local memory. This design preserves long-horizon consistency while adapting to new content, achieving linear-time scalability. Experiments demonstrate that VideoSSM substantially reduces error accumulation, motion drift, and content repetition, enabling minute-scale coherence and robust prompt-adaptive interactive generation. Future directions include integrating explicit multi-modal conditioning, incorporating camera-aware and geometric priors, and extending the framework to controllable long-form video editing.

\newpage

{
    \small
    \bibliographystyle{ieeenat_fullname}
    \bibliography{main}
}

% \input{docs/x_suppl}


\end{document}